\documentclass{article}
\usepackage[utf8]{inputenc}
\usepackage[margin=1.25in]{geometry}
\usepackage{makecell}
\usepackage{braket}
\usepackage{amsmath}
\usepackage{graphicx}
\usepackage{xcolor}
\usepackage{hyperref}
\usepackage{amssymb}
\usepackage{tikz}
\usepackage{authblk}
\usetikzlibrary{positioning}
\usetikzlibrary{shapes.geometric}
\usetikzlibrary{backgrounds}
\usepackage[sorting=none]{biblatex}
\title{Interaction Decompositions for Tensor Network Regression}

\author[1,2]{Ian Convy\thanks{ian\_convy@berkeley.edu}}
\author[1,2]{K. Birgitta Whaley}

\affil[1]{Department of Chemistry, University of California, Berkeley, CA 94720, USA}
\affil[2]{Berkeley Quantum Information and Computation Center, University of California, Berkeley, CA 94720, USA}

\date{}

\begin{document}

\maketitle

\begin{abstract}
    It is well known that tensor network regression models operate on an exponentially large feature space, but questions remain as to how effectively they are able to utilize this space. Using a polynomial featurization, we propose the \textit{interaction decomposition} as a tool that can assess the relative importance of different regressors as a function of their polynomial degree. We apply this decomposition to tensor ring and tree tensor network models trained on the MNIST and Fashion MNIST datasets, and find that up to 75\% of interaction degrees are contributing meaningfully to these models. We also introduce a new type of tensor network model that is explicitly trained on only a small subset of interaction degrees, and find that these models are able to match or even outperform the full models using only a fraction of the exponential feature space. This suggests that standard tensor network models utilize their polynomial regressors in an inefficient manner, with the lower degree terms being vastly under-utilized.
\end{abstract}

\section{Introduction}

Tensor network regression has emerged as a promising and active area of machine learning research, having achieved impressive results on common benchmark tasks such as the Movie 100K~\cite{Novikov_Trofimov_Oseledets_2016}, MNIST~\cite{Stoudenmire_Schwab_2016}\cite{Glasser_Pancotti_Cirac_2018}\cite{Stoudenmire_2018}\cite{Chen_Pan_Dong_2021}, and Fashion MNIST~\cite{Glasser_Pancotti_Cirac_2018}\cite{Stoudenmire_2018}\cite{Chen_Pan_Dong_2021} datasets. The effectiveness of these models can be attributed to the tensor-product transformation that is applied to the data features, which maps the original feature vector into an exponentially large vector space. By performing linear operations on this expanded feature space, tensor network models are able to generate regression outputs that are highly \textit{non-linear} functions of the original features.   

In most tensor network models, the tensor-product transformation is constructed from a set of vector-valued functions that each act on only a single data feature. The form of these functions is important to the operation of the model, as it determines how regression on the transformed space is related to regression on the original feature space. Conventional wisdom regarding the choice of these functions can be traced back to the parallel works of Stoudenmire and Schwab~\cite{Stoudenmire_Schwab_2016} and Novikov et al.~\cite{Novikov_Trofimov_Oseledets_2016}, who each proposed a different transformation scheme. The method from~\cite{Stoudenmire_Schwab_2016} was inspired by techniques in quantum many-body physics, and mapped each feature $x \in [0, 1]$ into the L2-normalized vector $[\cos(\frac{\pi}{2}x),\ \sin(\frac{\pi}{2}x)]$. The approach in~\cite{Novikov_Trofimov_Oseledets_2016}, by contrast, was motivated by a desire to characterize interactions within categorical (discrete) data, and therefore had each feature mapped to the vector $[1,\ x]$. The advantage of this latter mapping is that every element of the transformed feature space is a product of some subset of the original features, which makes the resulting regression output easier to interpret.

The purpose of this work is to quantitatively assess how well tensor network models are able to utilize the exponential feature space induced by their tensor-product transformations. We shall focus specifically on models which are built upon the $[1,\ x]$ featurization from Novikov et al.~\cite{Novikov_Trofimov_Oseledets_2016}, since this allows us to easily interpret different regions of the transformed space in terms of \textit{interactions} (products) between the original features. To this end, we introduce the \textit{interaction decomposition} of a tensor network model, which casts the regression output as the sum of terms which each contain all feature products of a fixed \textit{degree}. Here the degree of an interaction is defined as the number of features that are multiplied together, such that, e.g., interactions of degree three take the form $x_1x_2x_3$. By applying this decomposition to tensor network models that were trained on a given machine learning task, we can determine the importance of each interaction degree to the final output of the model. Furthermore, by implementing new models that regress on only a subset of degrees, we can assess whether the tensor network models are under-utilizing those interactions. 

The remainder of this paper has the following structure. Sec.~\ref{sec:tensor_network_regression} provides an overview of tensor network regression, starting with a review of tensor operations and ending with a description of the tensor ring and tree tensor network architectures that we used for our tests. In Sec.~\ref{sec:interaction_decomposition}, we describe the motivation and mechanics of the interaction decomposition, and then apply it to tensor network classifiers trained on the MNIST and Fashion MNIST datasets. From these tests, we find that some models utilize up to three-quarters of all interaction degrees generated by the tensor-product transformation, which collectively contain roughly $10^{19}$ different regressors. However, we also determine that the tensor network classifiers are vastly under-utilizing the lower-degree interactions, since separate models trained using only interactions less than, e.g., sixth degree are able to achieve classifications accuracies very near those of the full regression models. We discuss the implications of these results and directions for future work in Sec.~\ref{sec:discussion}. The \hyperref[sec:appendix]{Appendix} contains technical details about the procedure used to carry out the interaction decompositions, as a well as a tabulation of important numerical results.

\section{Tensor Network Regression}\label{sec:tensor_network_regression}

\subsection{Background}\label{sec:background}

\subsubsection{Tensor Overview}

Throughout this work, we consider machine learning models that are constructed using \textit{tensors}~\cite{Kolda_Bader_2009}\cite{Hackbusch_2012}. For our purposes, a tensor is simply a multidimensional array of numbers, such that each number is indexed by a non-negative integer along every dimension. The \textit{order} of a tensor is equal to the number of dimensions that it has, or equivalently the number of integers needed to specify one of its elements. From this perspective, vectors and matrices can be viewed as first-order and second-order tensors respectively. We will denote tensors with order greater than one using uppercase letters $(A, B, C, ...)$, while vectors will be denoted using a lower case letter under an arrow $(\vec{a}, \vec{b}, \vec{c}, ...)$. Elements of a tensor are specified using subscripts, so that an element of the third-order tensor $A$ is given by $A_{ijk}$, where $i, j, k$ are non-negative integers (all dimensions are indexed starting from zero). When referring to elements of a vector, the arrow symbol is dropped. To specify the $i$th member of a set of tensors, we use a superscript with parentheses, e.g., $A^{(i)}$. 

There are a wide variety of operations that can be defined between tensors, but in this work we focus primarily on the \textit{tensor product} and the \textit{tensor contraction}. The tensor product $C = A \otimes B$ constructs a new tensor $C$ from every pairwise product between elements of tensor $A$ and elements of tensor $B$, such that each element of $C$ is given by
\begin{equation}\label{eq:tensor_product}
    C_{i_0...i_{m-1}j_0...j_{n-1}} = A_{i_0...i_{m-1}}B_{j_0...j_{n-1}}
\end{equation}
and the order of $C$ is the sum of the orders of $A$ and $B$. The tensor contraction between $A$ and $B$ is similar to the corresponding tensor product, except that it generates a new tensor $C$ by taking elements of $A \otimes B$ and summing them along a set of specified dimensions. As an example, if $A$ and $B$ are both third order, then a contraction between the second dimension of $A$ and the third dimension of $B$ is written as
\begin{equation}\label{eq:tensor_contraction}
    C_{jklm} = \sum_{i}A_{jik}B_{lmi}.
\end{equation}
Note that $C$ is fourth order rather than sixth order, since two of the dimensions of $A \otimes B$ were summed together. 

\subsubsection{Tensor Networks}

Although we have described the tensor product and the tensor contraction as operations which construct a new tensor $C$ from existing tensors $A$ and $B$, it is equally valid to take the reverse view and interpret Eqs.~(\ref{eq:tensor_product},~\ref{eq:tensor_contraction}) as representing an existing tensor $C$ in terms of new components $A$ and $B$. This approach is the foundation of the \textit{tensor network} representation, in which a tensor of interest is decomposed into a set of contractions between component tensors~\cite{Biamonte_Bergholm_2017}\cite{Bridgeman_Chubb_2017}. The number and order of these component tensors are ultimately arbitrary, but most networks are constructed such that the component orders are all significantly smaller than the order of the original tensor. Since the number of elements in a tensor scales exponentially with its order (for a fixed index size), these component tensors will generally contain far fewer elements than the original tensor. This can allow for operations to be performed on the network that would have been computationally intractable on the original, higher-order tensor. 

As a very simple example of a tensor network, consider the sixth-order tensor $C$ that is represented by the contraction of two fourth-order tensors $A$ and $B$, such that the elements of $C$ are given by
\begin{equation}\label{eq:contraction}
    C_{ijklmn} = \sum^{t-1}_{r=0} A_{ijkr} B_{lmnr}.
\end{equation}
Assuming that each index in Eq.~\eqref{eq:contraction} is of size $t > 2$, it is clear that tensor $C$ has significantly more elements ($t^6$) than are contained in $A$ and $B$ combined ($2t^4$). Using this representation, operations on $C$ which are localized to indices $\{i,j,k\}$ or $\{l,m,n\}$ can instead be performed on $A$ or $B$ respectively, dramatically reducing their computational cost. Further inspection of Eq.~\eqref{eq:contraction} shows that the contraction of $A$ and $B$ can yield a combined tensor that has rank at most $t$ across dimension $r$, which is far smaller than the largest possible rank of $t^3$. This implies that most sixth-order tensors can only be approximately represented by the contraction of $A$ and $B$, with the quality of the approximation depending on both the underlying rank structure of $C$ and the size constraints placed on $A$ and $B$. Similar trade-offs occur across all tensor networks, with different contraction structures being better suited to represent different higher-order tensors~\cite{Grasedyck_Kressner_Tobler_2013}\cite{Evenbly_Vidal_2011}.

\begin{figure}
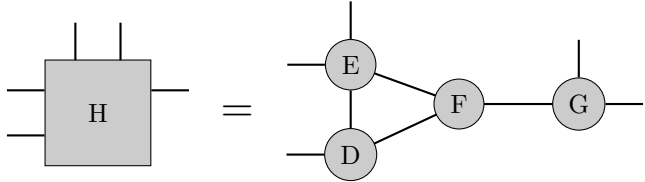

    \centering
    \begin{equation*}
        H_{ijkop} = \sum^{t-1}_{l,m,n,r = 0} D_{lim}E_{jmrk}F_{lrn}G_{nop}
    \end{equation*}
    \tikz[baseline=-6ex, tensor/.style={draw, fill=gray!40, minimum size=40}] {
    \node[tensor] (1) {H};    \node[] (text) [right=0.8 of 1] {\scalebox{1.6}{=}};
    \draw[black, thick] (-0.71, 0.30) -- (-1.21, 0.30);
    \draw[black, thick] (-0.71, -0.30) -- (-1.21, -0.30);
    \draw[black, thick] (0.71, 0.30) -- (1.21, 0.30);
    \draw[black, thick] (-0.30, 0.70) -- (-0.30, 1.20);
    \draw[black, thick] (0.30, 0.70) -- (0.30, 1.20);
    } \hspace{0.07cm}
    \tikz[tensor/.style={circle, draw, fill=gray!40, minimum size=15}]{
        \node[tensor] (1) {D};
        \node[tensor] (2) [above = 0.5 of 1.north] {E};
        \node[tensor] (3) [above right = 0.08 and 1.2 of 1.north] {F};
        \node[tensor] (4) [right = 0.9 of 3.east] {G};
        \begin{scope}[on background layer]
            \draw[black, thick] (1.west){} + (-0.5, 0) -- (1.west);
            \draw[black, thick] (1.north) -- (2.south);
            \draw[black, thick] (2.west){} + (-0.5, 0) -- (2.west);
            \draw[black, thick] (2.north){} + (0, 0.5) -- (2.north);
            \draw[black, thick] (2.center) -- (3.center);
            \draw[black, thick] (1.center) -- (3.center);
            \draw[black, thick] (3.center) -- (4.center);
            \draw[black, thick] (4.north){} + (0, 0.5) -- (4.north);
            \draw[black, thick] (4.east){} + (0.5, 0) -- (4.east);
        \end{scope}}
    \caption{Tensor network representing fifth-order tensor $H$, depicted using explicit summations (top) and a tensor diagram (bottom). The network is generated by four different contractions, each of which is indicated in the diagram by a shared leg. While it is possible to discern the contraction pattern of the component tensors by studying the index notation, the diagram makes it obvious at a glance.}
    \label{fig:diagram}
\end{figure}

Due to its simplicity, the network formed from $A$ and $B$ in Eq.~\eqref{eq:contraction} can be clearly conveyed by simply listing out all of the indices explicitly. However, many tensor networks involve contractions between dozens or even hundreds of tensors, each with their own set of indices. For notational clarity, it is common to use \textit{tensor diagrams} (also referred to as \textit{Penrose notation}~\cite{Penrose_1971}) to represent the sets of contractions within more complicated tensor networks. In these diagrams, each tensor is denoted using a geometric shape, while each index is represented by a line or \textit{leg} protruding outward from the shape. A tensor product is implied by placing two tensors next to one another, and a contraction is indicated by having those tensors share one or more legs. Figure~\ref{fig:diagram} shows the relative simplicity of the diagram notation versus explicit summations when several tensors are involved in the network. We utilize tensor diagrams, along with explicit index expressions, throughout the remainder of this paper to help illustrate the relevant tensor operations.

\subsubsection{Regression with Tensors}\label{sec:tensor_regression}

Our work focuses on the use of tensor networks as a means of performing \textit{regression}~\cite{Hastie_Tibshirani_Friedman_2009}. In a regression task, the goal is to learn (or estimate) the relationship between a set of $m$ independent variables $\{x_i\}^{m-1}_{i=0}$ called \textit{features} and a set of $n$ dependent variables $\{y_i\}^{n-1}_{i=0}$ called \textit{labels}. We denote a joint sample of these $m+n$ variables as $(\vec{x}, \vec{y})$, where $\vec{x} \in \mathbb{R}^m$ is a vector containing the values of the features and $\vec{y} \in \mathbb{R}^n$ is a vector containing the values of the labels. In the case of \textit{parametric} regression, the relationship between $\vec{x}$ and $\vec{y}$ is modeled by a function $\vec{f}$ such that
\begin{equation}\label{eq:regression}
\vec{y} \approx \vec{f}(\vec{x}; \mathcal{W}),
\end{equation}
where $\mathcal{W}$ is a set of learned parameters which determines the behavior of the function. We do not expect the relationship in Eq.~\eqref{eq:regression} to be exact for any intuitive function $\vec{f}$, except when the data is generated artificially. Indeed, an exact reconstruction will generally be undesirable for real-world data, since the labels are likely to contain noise that should not be directly copied into the model. Once $\vec{f}$ is learned, it can be used to make predictions about the value of $\vec{y}$ for an unlabeled sample $\vec{x}$.

Tensor network regression can be understood as a specific form of \textit{tensor regression}~\cite{Liu_Liu_Long_Zhu_2022}, in which a large tensor of regression coefficients is generated by a network of smaller component tensors. In tensor regression, the function $\vec{f}$ of Eq.~\eqref{eq:regression} is expressed as the contraction of a data tensor $X(\vec{x})$, which is a function of the features in a given sample, and a weight tensor $W$ whose elements make up the set of parameters $\mathcal{W}$. The data tensor can, in principle, take on any form, but it is usually constructed from the tensor product of a set of $m$ vector-valued functions $\{\vec{h}^{(i)}\}^{m-1}_{i=0}$ that each take as input a single feature:
\begin{equation}\label{eq:data_tensor}
    X(\vec{x}) = \bigotimes^{m-1}_{i=0}\vec{h}^{(i)}(x_i) \enspace \rightarrow \enspace \tikz[baseline=0.1ex, data/.style={draw, fill=gray!40, minimum size=7}] {
        \node[data] (1) {};
        \draw[black, thick] (1.north){}+(0, 0.25) -- (1.north);
        \foreach \x [count=\i] in {2,3} {
            \node[data] (\x) [right=0.25 of \i] {};
            \draw[black, thick] (\x.north){}+(0, 0.25) -- (\x.north);
        }
}\enspace ... \enspace \tikz[baseline=0.1ex, data/.style={draw, fill=gray!40, minimum size=7}] {
        \node[data] (4) {};
        \draw[black, thick] (4.north){}+(0, 0.25) -- (4.north);
} \enspace,  
\end{equation}
where $x_i$ is the $i$th element of $\vec{x}$ and thus the $i$th feature out of $m$. Note that the sequence of tensor products in Eq.~\eqref{eq:data_tensor} is similar to a tensor network, insofar as it expresses a higher-order tensor $X$ using a set of order-one components which collectively contain exponentially fewer elements. The regression output $\vec{f}(\vec{x};\mathcal{W})$ is computed by contracting the weight tensor $W$ with $X$:
\begin{equation}\label{eq:tensor_regression}
    f_k(\vec{x};\mathcal{W}) = \sum_{i_0...i_{m-1}} W_{ki_0...i_{m-1}}X_{i_0...i_{m-1}}(\vec{x}) \enspace \rightarrow \enspace 
    \tikz[baseline=1.3ex, data/.style={draw, fill=gray!40, minimum size=7}] {
        \node[data] (1) {};
        \draw[black, thick] (1.north){}+(0, 0.25) -- (1.north);
        \foreach \x [count=\i] in {2,3} {
            \node[data] (\x) [right=0.25 of \i] {};
            \draw[black, thick] (\x.north){}+(0, 0.25) -- (\x.north);
        }
        \node[] (text) [right=0.1 of 3] {...};
        \node[data] (4) [right=0.1 of text] {};
        \draw[black, thick] (4.north){}+(0, 0.25) -- (4.north);
        \node[draw, fill=gray!40, minimum height = 13, minimum width = 64.5] (weights) [above right = 0.25 and -0.25 of 1] {\textit{W}};
        \draw[black, thick] (weights.north){}+(0, 0.25) -- (weights.north);
        }\enspace ,
\end{equation}
where $W$ contains an additional dimension $k$ that indexes the output vector of the model. This form of regression is quite distinct from the standard
deep learning paradigm, in that the transformation of the data is effectively set in
advance here via the data tensor featurization $X(\vec{x})$. All that is then left to optimize are the coefficients $W_{ki_0...i_m}$ that should be assigned to each of the new regressors. The same delineation cannot in general be made
for deep learning models, since they are formed from a composition of non-linear functions that has
no clear relation to any series expansion.

The precise role that the original features  $\{x_i\}^{m-1}_{i=0}$ play in Eq.~\eqref{eq:tensor_regression} depends on the form of $X$ and therefore on the set of functions $\{\vec{h}^{(i)}\}^{m-1}_{i=0}$ that were chosen. For our work we follow Novikov et al. and use functions of the form
\begin{equation}\label{eq:feat_func}
    \vec{h}^{(i)}(x_i) = \begin{bmatrix} 1 \\ x_i \end{bmatrix},
\end{equation}
which have been used in other implementations of tensor network regression~\cite{Stoudenmire_2018}\cite{Efthymiou_Hidary_Leichenauer_2019}\cite{Meng_Zhang_Zhang_Gao_Ran_2021}. When Eq.~\eqref{eq:feat_func} is used to construct $X$, the regression function $\vec{f}(\vec{x};\mathcal{W})$ from Eq.~\eqref{eq:tensor_regression} becomes
\begin{equation}\label{eq:tensor_regression_pow}
    f_k(\vec{x};\mathcal{W}) = \sum^1_{i_0...i_{m-1} = 0} W_{ki_0...i_{m-1}}x^{i_0}_0x^{i_1}_1\cdots x^{i_{m-1}}_{m-1},
\end{equation}
where $0^0 = 1$ is assumed. Eq.~\eqref{eq:tensor_regression_pow} shows that tensor regression, when using the definition of $\vec{h}^{(i)}(x_i)$ from Eq.~\eqref{eq:feat_func}, is equivalent to linear regression on all possible products formed between the original features, plus a bias term when all of the indices are zero. Regression on the elements of $X$ can therefore generate functions $\vec{f}$ which have non-zero mixed derivatives with respect to the original features. For example,
\begin{equation}
    \frac{\partial^2}{\partial x_0\partial x_1} f_k(\vec{x};\mathcal{W}) = \sum^1_{i_2...i_{m-1} = 0} W_{k11i_2...i_{m-1}}x^{i_2}_2x^{i_3}_3\cdots x^{i_{m-1}}_{m-1}.
\end{equation}
These non-zero derivatives make $\vec{f}$ significantly more expressive than functions generated by linear regression directly on the original features in $\vec{x}$, for which all mixed derivatives must vanish.

\subsection{Regression using Tensor Rings and Tree Tensor Networks}\label{sec:tr_ttn_regression}

Although straightforward mathematically, the approach to tensor regression outlined in Sec.~\ref{sec:tensor_regression} is often impractical due to the size of the weight tensor $W$. 
Inspection of Eq.~\eqref{eq:tensor_regression_pow} reveals that there are $2^m$ parameters in $W$, which is exponential in the number of features. Given that many regression tasks involve data with hundreds or even thousands of features, this method of parameterization cannot be used. Tensor network regression offers an alternative approach, in which the weight tensor is decomposed into a set of low-order component tensors. The parameters $\mathcal{W}$ of the regression model are then taken to be the elements of these tensors, rather than the elements of $W$ directly. This decomposition can be illustrated using Figure~\ref{fig:diagram}, in which $H$ serves as the weight tensor while the components $D$, $E$, $F$, and $G$ contain the actual parameters of the model. Since the data tensor $X$ is itself composed of vectors $\{\vec{h}^{(i)}(x_i)\}^{m-1}_{i=0}$, it is possible for the contraction in Eq.~\eqref{eq:tensor_regression} to be carried out using only the \textit{components} of $W$ and $X$. 

\begin{figure}
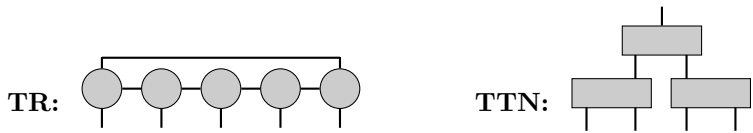

    \centering
   \textbf{ TR:} \enspace \tikz[baseline=-2ex, mps/.style={draw, circle, fill=gray!40, minimum size=15}, data/.style={draw, fill=gray!40, minimum size=10}] {
        \node[mps] (1) {};
        \node[mps] (2) [right=0.25 of 1] {};
        \node[mps] (3) [right=0.25 of 2] {};
        \node[mps] (4) [right=0.25 of 3] {};
        \node[mps] (5) [right=0.25 of 4] {};
        \draw[black, thick] (1.south){}+(0, -0.25) -- (1.south);
        \draw[black, thick] (1.north){}+(0, 0.15) -- (1.north);
        \draw[black, thick] (2.south){}+(0, -0.25) -- (2.south);
        \draw[black, thick] (1.east) -- (2.west);
        \draw[black, thick] (4.south){}+(0, -0.25) -- (4.south);
        \draw[black, thick] (2.east) -- (3.west);
        \draw[black, thick] (3.south){}+(0, -0.25) -- (3.south);
        \draw[black, thick] (5.south){}+(0, -0.25) -- (5.south);
        \draw[black, thick] (3.east) -- (4.west);
        \draw[black, thick] (4.east) -- (5.west);
        \draw[black, thick] (5.north){}+(0, 0.15) -- (5.north);
        \draw[black, thick] ([yshift=4] 1.north) -- ([yshift=4] 5.north);
        }
    \quad \quad \quad \quad
    \textbf{TTN:} \enspace 
    \tikz[baseline=-1.6ex, data/.style={draw, fill=gray!40, minimum size=10}]{
    \node[draw, anchor=west, fill=gray!40, minimum height= 11, minimum width=30] (1_left) {};
    \draw[black, thick] (0.20,-0.19) -- (0.20,-0.50);
    \draw[black, thick] (0.85,-0.19) -- (0.85,-0.50);
    
    \node[draw, fill=gray!40, minimum height= 11, minimum width=30] (2_left) [right=0.25 of 1_left] {};
    \draw[black, thick] (1.53,-0.19) -- (1.53,-0.50);
    \draw[black, thick] (2.18,-0.19) -- (2.18,-0.50);
    
    \node[draw, fill=gray!40, minimum height= 11, minimum width=30] (upper) [above right=0.30 and -0.40 of 1_left] {};
    \draw[black, thick] (0.85,0.51) -- (0.85,0.19);
    \draw[black, thick] (1.53,0.51) -- (1.53,0.19);
    \draw[black, thick] (upper.north){}+(0,0.25) -- (upper.north);
    }
    \caption{Tensor diagrams depicting a tensor ring (TR) and a tree tensor network (TTN), which can both be used for regression. The five component tensors of the TR are arranged in a line and connected together horizontally by five virtual indices, with a physical index dangling vertically from each component. By contrast, the three components of the TTN are arranged in a binary tree, with a pair of virtual indices connecting the two layers. The physical indices of the TTN are placed along the bottom layer, although when used for regression an additional physical index is added to the top tensor to represent the output, as shown here.}
    \label{fig:mps_ttn}
\end{figure}

There exist a wide variety of tensor network architectures which can be used for regression. Our work here will focus on two of the more popular designs: tensor rings~\cite{Zhao_Zhou_Xie_Zhang_Cichocki_2016}\cite{Mickelin_Karaman_2020} and tree tensor networks~\cite{Shi_Duan_Vidal_2006}\cite{Oseledets_Tyrtyshnikov_2009}. The structures of these networks are illustrated using tensor diagrams in Figure~\ref{fig:mps_ttn}, and are described in the following subsections.

\subsubsection{Tensor Rings}\label{sec:mps_regression}

The tensor ring (TR) is a popular type of tensor network, having been utilized for neural network compression~\cite{Wang_Sun_Eriksson_Wang_Aggarwal_2018}\cite{Pan_Xu_Wang_Ye_Wang_Bai_Xu_2019} and image reconstruction~\cite{Yuan_Li_Mandic_Cao_Zhao_2019}\cite{Zhao_Sugiyama_Yuan_Cichocki_2019}\cite{He_Yokoya_Yuan_Zhao_2019}. As depicted in Figure~\ref{fig:mps_ttn}, a TR is constructed from a \mbox{1-D} sequence of third-order tensors contracted along a set of \textit{virtual indices} that link neighboring tensors together. The ``ring'' part of a TR references the fact that the tensors at the beginning and end of the sequence are also contracted together, forming a closed loop. In addition to its two virtual indices, each component tensor also has a \textit{physical index}, which becomes an index of the higher-order tensor after contraction of the virtual indices. When a TR is used for regression, the physical indices are contracted with the $\{\vec{h}^{(i)}(x_i)\}^{m-1}_{i=0}$ from the data tensor, except for one tensor whose physical index is left uncontracted to serve as the index of $\vec{f}$. The TR is closely related to the matrix product state (MPS), which is used heavily in quantum many-body physics~\cite{Schollwöck_2011} and also frequently utilized for tensor network regression
\cite{Novikov_Trofimov_Oseledets_2016}\cite{Stoudenmire_Schwab_2016}\cite{Efthymiou_Hidary_Leichenauer_2019}. The structure of an MPS network, also referred to as a \textit{tensor train decomposition}~\cite{Oseledets_2011}, is identical to that of a TR, except that the tensors at the ends of the chain are second-order and thus not contracted together. In this work we use TRs rather than MPSs due to the greater symmetry of the former, which allows us to employ simpler contraction algorithms.

For tensor network regression to be practical, the sequence of contractions between components of $X$ and components of the network must be carried out such that all of the intermediate tensors are low-order. For a TR, this can be easily achieved by performing the contractions in two stages, with the vectors $\{\vec{h}^{(i)}(x_i)\}^{m-1}_{i=0}$ first being contracted with their corresponding component tensors in the TR along the physical index, which produces a new set of second-order tensors. These matrices are then contracted together, along with the third-order tensor containing the regression output, using the virtual indices. For $m = 4$, the diagrams of these steps are given by
\begin{equation}\label{eq:mps_regression}
    \tikz[baseline=-2ex, mps/.style={draw, circle, fill=gray!40, minimum size=7}, data/.style={draw, fill=gray!40, minimum size=7}] {
        \node[mps] (1) {};
        \node[mps] (2) [right=0.25 of 1] {};
        \node[mps] (3) [right=0.25 of 2] {};
        \node[mps] (4) [right=0.25 of 3] {};
        \node[mps] (5) [right=0.25 of 4] {};
        \node[data] (1_data) [below=0.25 of 1] {};
        \node[data] (2_data) [below=0.25 of 2] {};
        \node[data] (3_data) [below=0.25 of 4] {};
        \node[data] (4_data) [below=0.25 of 5] {};
        \draw[red, thick] (1.south){}+(0, -0.25) -- (1.south);
        \draw[black, thick] (1.north){}+(0, 0.15) -- (1.north);
        \draw[red, thick] (2.south){}+(0, -0.25) -- (2.south);
        \draw[black, thick] (1.east) -- (2.west);
        \draw[red, thick] (4.south){}+(0, -0.25) -- (4.south);
        \draw[black, thick] (2.east) -- (3.west);
        \draw[black, thick] (3.south){}+(0, -0.25) -- (3.south);
        \draw[red, thick] (5.south){}+(0, -0.25) -- (5.south);
        \draw[black, thick] (3.east) -- (4.west);
        \draw[black, thick] (4.east) -- (5.west);
        \draw[black, thick] (5.north){}+(0, 0.15) -- (5.north);
        \draw[black, thick] ([yshift=4] 1.north) -- ([yshift=4] 5.north);
        }
        \hspace{0.5ex} \rightarrow \hspace{0.7ex}
         \tikz[baseline=-1ex, matrix/.style={draw, circle, fill=gray!80, minimum size=7}]{
            \node[matrix] (1) {};
            \node[matrix] (2) [right=0.25 of 1] {};
            \node[draw, circle, fill=gray!40, minimum size=7] (3) [right=0.25 of 2] {};
            \node[matrix] (4) [right=0.25 of 3] {};
            \node[matrix] (5) [right=0.25 of 4] {};
            \draw[red, thick] (1) -- (2);
            \draw[red, thick] (2) -- (3);
            \draw[red, thick] (3) -- (4);
            \draw[red, thick] (4) -- (5);
            \draw[black, thick] (3.south){}+(0, -0.25) -- (3.south);
            \draw[red, thick] ([yshift=4] 1.north) -- (1.north);
            \draw[red, thick] ([yshift=4] 5.north) -- (5.north);
            \draw[red, thick] ([yshift=4] 1.north) -- ([yshift=4] 5.north);
        }
        \hspace{0.5ex} \rightarrow \hspace{0.5ex}
        \tikz[baseline=-1ex] {
            \node[draw, circle, fill=gray!80, minimum size=7] (1) {}; 
            \draw[black, thick] (1.south){}+(0, -0.25) -- (1.south);
        }
        \hspace{0.5ex} = \hspace{0.5ex}
        \vec{f}(\vec{x};\mathcal{W})
\end{equation}
where the legs colored in red are those that are contracted from one step to the next. When using the functions from Eq.~\eqref{eq:feat_func}, the total number of parameters in a TR regression model is $2(m+1)r^2$, where $r$ is the size of the virtual indices. Since the number of features $m$ is generally fixed for a given regression task (after preprocessing), $r$ serves as the primary tunable parameter in the model, with larger values of $r$ placing fewer restrictions on the elements of $W$. If $r$ is allowed to grow exponentially with $m$, then the TR can represent an arbitrary weight tensor $W$, although this generally defeats the purpose of using a tensor network. In practice, $r$ is typically capped at between $10\text{ -- }100$ for regression.

\subsubsection{Tree Tensor Networks}\label{sec:ttn_regression}

A common alternative to the TR/MPS network is the tree tensor network (TTN), in which the component tensors are arranged in a (typically binary) tree pattern. TTNs have been used for quantum simulation~\cite{Shi_Duan_Vidal_2006}\cite{Murg_Verstraete_Legeza_Noack_2010}, efficient tensor representation~\cite{Grasedyck_2010} (where it is known as the \textit{hierarchical Tucker decomposition}), and for regression~\cite{Liu_Ran_Wittek_Peng_Garcia_Su_Lewenstein_2019}\cite{Stoudenmire_2018}. An example TTN is depicted in Figure~\ref{fig:mps_ttn}, showing that each component tensor has both a horizontal and vertical position in the network. Similar to a TR, a TTN contains both virtual and physical indices, but only the lowest layer of  component tensors are contracted directly with the data tensor $X$. While the structure of a TR can be applied easily to any value of $m$, a binary TTN works most efficiently when $m = 2^l$, where $l$ is the number of layers in the tree. Although these networks can be modified to handle other values of $m$, our work here will only consider regression tasks where the number of features is a power of two.

When used for tensor regression, the components in the bottom layer of the TTN are first contracted with the $\{\vec{h}^{(i)}(x_i)\}^{m-1}_{i=0}$ of $X$, which generates a new layer of $\frac{m}{2}$ first-order tensors. This is shown in the second diagram of Eq.~\eqref{eq:ttn_regression}, where the new first-order tensors are depicted as dark blocks. These tensors are then contracted with the second layer of the tree, generating $\frac{m}{4}$ first-order tensors. This process repeats layer-by-layer until the regression output $\vec{f}(\vec{x})$ is generated by contracting the top tensor, which is shown in the third diagram of Eq.~\eqref{eq:ttn_regression}. The intermediate tensors created at each layer are always first-order, which ensures that the procedure will be computationally tractable. For $m = 4$, the tensor diagrams for the contractions are given by
\begin{equation}\label{eq:ttn_regression}
\tikz[baseline=-1.6ex, data/.style={draw, fill=gray!40, minimum size=7}]{
    \node[draw, anchor=west, fill=gray!40, minimum height= 7, minimum width=20] (1_left) {};
    \draw[red, thick] (0.13,-0.12) -- (0.13,-0.37);
    \draw[red, thick] (0.6,-0.12) -- (0.6,-0.37);
    \node[data] (1_data) [below right = 0.24 and -0.72 of 1_left] {};
    \node[data] (1_data) [below right = 0.24 and -0.25 of 1_left] {};
    
    \node[draw, fill=gray!40, minimum height= 7, minimum width=20] (2_left) [right=0.25 of 1_left] {};
    \draw[red, thick] (1.09,-0.12) -- (1.09,-0.37);
    \draw[red, thick] (1.56,-0.12) -- (1.56,-0.37);
    \node[data] (1_data) [below right = 0.24 and 0.24 of 1_left] {};
    \node[data] (1_data) [below right = 0.24 and 0.71 of 1_left] {};
    
    \node[draw, fill=gray!40, minimum height= 7, minimum width=20] (upper) [above right=0.25 and -0.23 of 1_left] {};
    \draw[black, thick] (0.6,0.38) -- (0.6,0.13);
    \draw[black, thick] (1.09,0.38) -- (1.09,0.13);
    \draw[black, thick] (upper.north){}+(0,0.25) -- (upper.north);
    }
    \hspace{0.5ex} \rightarrow \hspace{0.5ex} 
    \tikz[baseline=0ex]{
    \node[draw, anchor=west, fill=gray!80, minimum height= 7, minimum width=20] (1_left) {};
    \node[draw, fill=gray!80, minimum height= 7, minimum width=20] (2_left) [right=0.25 of 1_left] {};
    \node[draw, fill=gray!40, minimum height= 7, minimum width=20] (upper) [above right=0.25 and -0.23 of 1_left] {};
    \draw[red, thick] (0.6,0.38) -- (0.6,0.13);
    \draw[red, thick] (1.09,0.38) -- (1.09,0.13);
    \draw[black, thick] (upper.north){}+(0,0.25) -- (upper.north);
    }
    \hspace{0.5ex} \rightarrow \hspace{0.5ex}
    \tikz[]{\node[draw, fill=gray!80, minimum height= 7, minimum width=20] (upper){};
    \draw[black, thick] (upper.north){}+(0,0.25) -- (upper.north);
    }
    \hspace{0.5ex} = \hspace{0.5ex}
    \vec{f}(\vec{x};\mathcal{W}),
\end{equation}
with the number of tensors being approximately halved after each layer is contracted. As in Eq.~\eqref{eq:mps_regression}, the legs shown in red are contracted between each step. When using $\vec{h}^{(i)}(x_i)$ of the form in Eq.~\eqref{eq:feat_func}, the number of parameters in a TTN is $2mr + (\frac{m}{2}-1)r^3$, which scales as $\mathcal{O}(r^3)$ in contrast with the $\mathcal{O}(r^2)$ scaling of a TR. As a result, the size $r$ of the virtual index is typically chosen to be on the order of 10. As with a TR, a TTN can represent an arbitrary weight tensor $W$ if $r$ is allowed to scale exponentially with $m$.

\section{Interaction Decomposition}\label{sec:interaction_decomposition}

\subsection{Motivation}

Throughout our discussion of tensor network regression in Sec.~\ref{sec:tr_ttn_regression}, the weight tensor $W$ and data tensor $X$ were treated principally as abstract objects, in that they were only operated on numerically via their component tensors. This was necessary on practical grounds, since the exponential scaling of both $W$ and $X$ makes it virtually impossible to perform operations on either tensor when the data has even a modest number of features $m$. That said, there is an obvious mathematical clarity that comes from working directly with $W$ and $X$ via the decomposition of Eq.~\eqref{eq:tensor_regression_pow}, since the elements of $X$ are simply products of the original features while the elements of $W$ are the corresponding linear regression coefficients. If, for example, we wished to perform regression using only a specific portion of the feature products, then we could just set the elements of $W$ for all other feature products to zero and learn the remaining parameters as usual. Such a straightforward modification is generally not possible when representing the weight tensor as a tensor network, since each element of $W$ is a complicated function of all of the parameters in the model.

In this section we introduce the \textit{interaction decomposition} of a tensor network, with the aim of recovering some of the fine-tuned control and interpretability that comes from an element-wise representation of the weight tensor $W$. In an interaction decomposition, the terms of the sum in Eq.~\eqref{eq:tensor_regression_pow} are grouped together by the number of features included in their product, for a total of $m + 1$ groupings. The number of features in a given product is labeled its interaction \textit{degree}, such that $x_1$ has degree 1, $x_1x_2$ has degree 2, and so on, with the bias having degree 0. Under an interaction decomposition, the regression output $\vec{f}(\vec{x};\mathcal{W})$ is written as
\begin{equation}\label{eq:degree_decomp}
    \vec{f}(\vec{x};\mathcal{W}) = \sum^m_{j=0} \vec{d}^{\ (j)}(\vec{x};\mathcal{W}),
\end{equation}
where $\vec{d}^{\ (j)}(\vec{x};\mathcal{W})$ is the contribution to the regression output from all terms of degree $j$. As with $\vec{f}(\vec{x};\mathcal{W})$, these contributions are functions of both the original features $\vec{x}$ and the parameters $\mathcal{W}$ of the decomposed network. We discuss ways of interpreting this decomposition in terms of vector subspaces in Sec.~\ref{sec:degree_subspaces}.

Using Eq.~\eqref{eq:degree_decomp}, the relative importance of the $j$th interaction degree can be assessed by analyzing the average magnitude of $\vec{d}^{\ (j)}(\vec{x};\mathcal{W})$, as well as its effect on the regression output. We carry out this analysis on TR and TTN models in Sec.~\ref{sec:tr_ttn_decomp}. Furthermore, by choosing to keep only a specific subset $\mathcal{D}$ of the decomposition terms in Eq.~\eqref{eq:degree_decomp}, it is possible to construct a new type of regression model which we call a $\mathcal{D}$\textit{-degree tensor network}. These networks utilize the same parameterization scheme for $W$ as normal tensor network models of the same architecture, but are restricted to generating only the feature products of degrees contained in $\mathcal{D}$. By comparing the performance of a full tensor network with that of a $\mathcal{D}$-degree version of the network, we can quantify how effectively the standard network is able to utilize interaction degrees within $\mathcal{D}$. We introduce these models and perform numerical tests on them in Sec.~\ref{sec:decomp_regression}.

\subsection{Interaction Subspaces}\label{sec:degree_subspaces}

In the context of vector spaces, the weight tensor $W$ acts as a parameterized multilinear map between the data tensor space $\mathbb{X}$, which we take to be $\mathbb{R}^{2^{m}}$, and the label space $\mathbb{Y}$. Under this construction, the data tensor $X$ is simply a vector within $\mathbb{X}$ whose elements are generated from the features in $\vec{x}$ via the tensor-product operations of Eq.~\eqref{eq:data_tensor}. The nature of the $\vec{d}^{\ (j)}(\vec{x};\mathcal{W})$ terms from Eq.~\eqref{eq:degree_decomp} can be understood by considering the action of $W$ on a particular subspace decomposition of $X$. Using the definition of $\vec{h}^{(i)}(x_i)$ from Eq.~\eqref{eq:feat_func}, we can expand $X$ on a standard basis of $\mathbb{X}$ as
\begin{equation}\label{eq:data_tensor_basis}
    X = \bigotimes^{m-1}_{i=0} (\vec{e}^{\ (i)}_0 + x_i\vec{e}^{\ (i)}_1)  = \sum^1_{i_1,...,i_m=0} x_0^{i_0}\cdots x_{m-1}^{i_{m-1}} \ \vec{e}^{\ (0)}_{i_0} \otimes ... \otimes \vec{e}^{\ (m-1)}_{i_{m-1}},
\end{equation}
where $\{\vec{e}^{\ (i)}_0, \vec{e}^{\ (i)}_1\}$ spans the two-dimensional space inhabited by $\vec{h}^{(i)}(x_i)$. In words, Eq.~\eqref{eq:data_tensor_basis} shows that the tensor products $\vec{e}^{\ (0)}_{i_0} \otimes ... \otimes \vec{e}^{\ (m-1)}_{i_{m-1}}$ form a basis for $\mathbb{X}$ upon which the coefficients of $X$ take the form of feature products.

Looking at the structure of the tensor-product basis used in Eq.~\eqref{eq:data_tensor_basis}, it is possible to divide $\mathbb{X}$ into the direct sum of subspaces $\{\mathbb{D}^{(0)}, ..., \mathbb{D}^{(m)}\}$, where $\mathbb{D}^{(j)}$ is a subspace spanned by the basis tensors $\vec{e}^{\ (0)}_{i_0} \otimes ... \otimes \vec{e}^{\ (m-1)}_{i_{m-1}}$ such that $j$ of the component basis vectors are of the form $e^{\ (i)}_1$ and $m-j$ are of the form $\vec{e}^{\ (i)}_0$. The coefficients of $X$ on the bases in $\mathbb{D}^{(j)}$ consist of feature products of degree $j$, so we therefore refer to $\mathbb{D}^{(j)}$ as the \textit{degree-$j$ subspace} of $\mathbb{X}$. The dimension of $\mathbb{D}^{(j)}$ is given by
\begin{equation}\label{eq:degree_dim}
    \text{dim}(\mathbb{D}^{(j)}) = \binom{m}{j},
\end{equation}
which is the number of ways to draw $j$ features from the total set of $m$ features. The dimension of the combined feature space for a set $\mathcal{D}$ of degrees, denoted $\dim(\mathcal{D})$, is then
\begin{equation}\label{eq:D_dim}
\text{dim}(\mathcal{D}) = \sum_{j \in \mathcal{D}} \binom{m}{j},
\end{equation}
which is simply the sum of the subspace dimension for each degree in $\mathcal{D}$. If we denote the projection into the degree-$j$ subspace as $P^{(j)}$, then the interaction decomposition becomes
\begin{equation}\label{eq:degree_decomp_proj}
    f_k(\vec{x};\mathcal{W}) = \sum^m_{j=0}\ \sum^1_{i_0,...,i_{m-1}=0}W_{ki_0...i_{m-1}}(P^{(j)}X)_{i_0...i_{m-1}} \rightarrow \sum^m_{j=0} \enspace
    \tikz[baseline=5ex, data/.style={draw, fill=gray!40, minimum size=7}] {
        \node[data] (1) {};
        \draw[black, thick] (1.north){}+(0, 1.1) -- (1.north);
        \foreach \x [count=\i] in {2,3} {
            \node[data] (\x) [right=0.25 of \i] {};
            \draw[black, thick] (\x.north){}+(0, 1.1) -- (\x.north);
        }
        \node[] (text) [right=0.1 of 3] {...};
        \node[data] (4) [right=0.1 of text] {};
        \draw[black, thick] (4.north){}+(0, 1.1) -- (4.north);
        \node[draw, fill=gray!40, minimum height = 13, minimum width = 64.5] (proj) [above right = 0.25 and -0.25 of 1] {$P^{(j)}$};
        \node[draw, fill=gray!40, minimum height = 13, minimum width = 64.5] (weights) [above right = 1.05 and -0.25 of 1] {\textit{W}};
        \draw[black, thick] (weights.north){}+(0, 0.25) -- (weights.north);
        }\enspace ,
\end{equation}
where the regression output is a sum of contractions between $W$ and the projection of $X$ into each of the $m + 1$ degree subspaces. Due to the linearity of tensor contractions, this equality can be easily verified by performing the sum over $j$ first, since $\sum_j P^{(j)}$ gives the identity.  

The form of Eq.~\eqref{eq:degree_decomp_proj} provides two interpretations of the degree contributions $\vec{d}^{\ (j)}(\vec{x};\mathcal{W})$. If we consider the projector $P^{(j)}$ acting on $X$, as originally envisioned, then $\vec{d}^{\ (j)}(\vec{x};\mathcal{W})$ is the contraction of $W$ with the portion of $X$ that inhabits $\mathbb{D}^{(j)}$. This could be viewed as a form of data tensor preprocessing, where elements of $X$ corresponding to interaction degrees other than $j$ are removed. Alternatively, the tensor diagrams in Eq.~\eqref{eq:degree_decomp_proj} show that it is equally valid to consider $P^{(j)}$ acting on the weight tensor $W$. Under this interpretation, the set $\{\vec{d}^{\ (j)}(\vec{x};\mathcal{W})\}^m_{j=0}$ consists of regressions on $X$ performed by different models, each derived from a common tensor $W$ by keeping only those elements corresponding to interactions of degree $j$. We will shift between these two interpretations freely throughout the remainder of this section, describing the interaction decomposition as a procedure which picks out different pieces of a tensor network model and which restricts the set of feature products that can be used for regression.

\subsection{Interaction Decompositions of TR and TTN Models}\label{sec:tr_ttn_decomp}

\begin{figure}
    \centering
    \includegraphics[width=\textwidth]{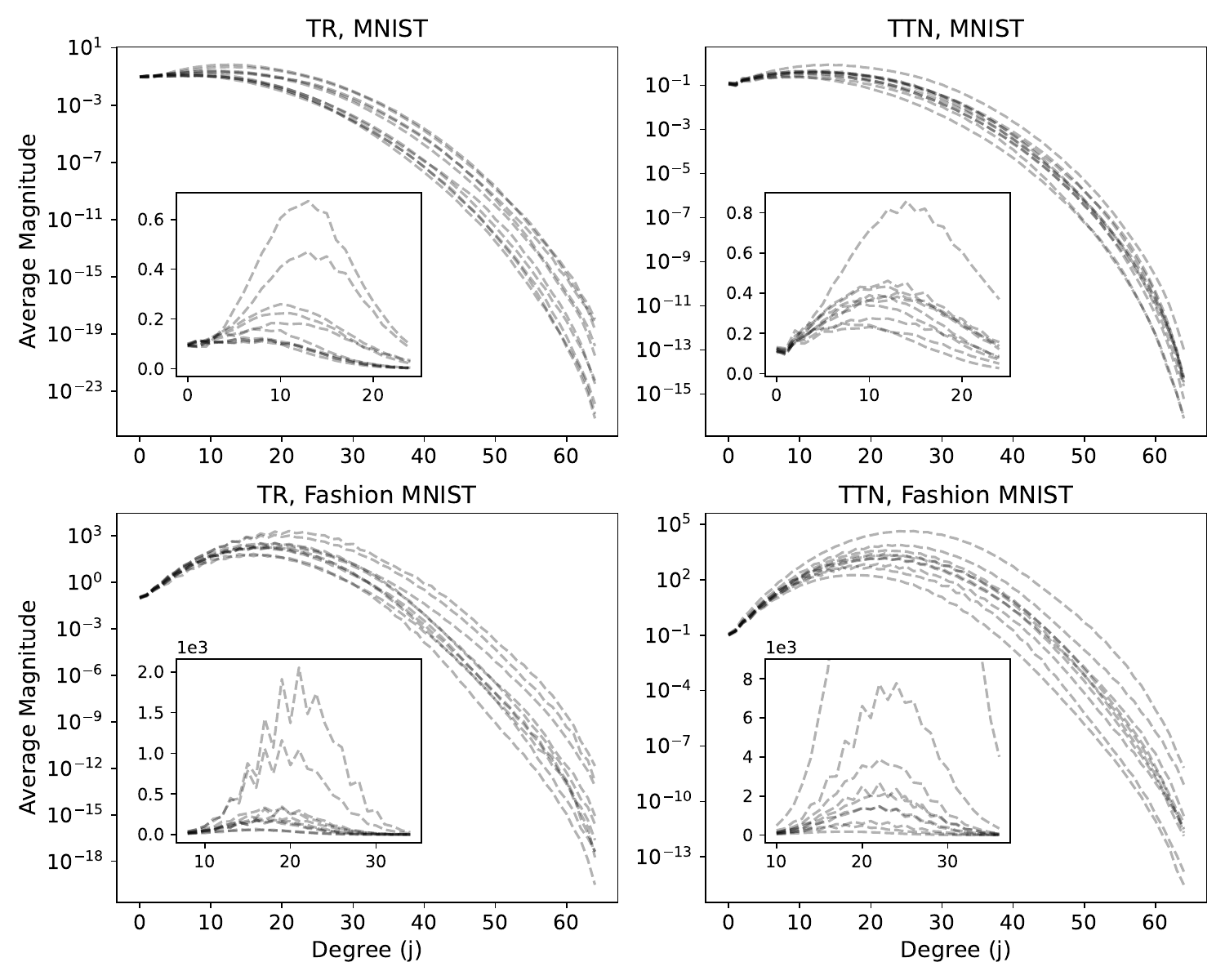}
    \caption{Plots of the L1 norm of $\vec{d}^{\ (j)}(\vec{x};\mathcal{W})$, averaged across the MNIST and Fashion MNIST test datasets, for ten TR models and ten TTN models trained using all interaction degrees. Each dashed line represents the average magnitudes from one of the models, which is plotted against the interaction degree $j$ of the contribution. The trend for all models is broadly the same, with magnitudes starting near $10^{-1}$ for the bias term and then gradually rising to a peak before dropping off significantly by degree 45. Large variations in magnitude can be seen when comparing individual TR models and TTN models.  
    }
    \label{fig:order_mag}
\end{figure}

The interaction decomposition of a TR or TTN regression model allows us to quantify the relative importance of the $j$th interaction degree to the overall value of the output. This information is not available when performing the standard contraction operations laid out in Eqs.~(\ref{eq:mps_regression}, \ref{eq:ttn_regression}), since the elements of the intermediate tensors are sums of contributions from a wide range of interaction degrees, making it impossible to separate out the impact of any one degree. Given the unfavorable scaling of Eq.~\eqref{eq:degree_dim}, a brute force evaluation of each feature product in $\vec{d}^{\ (j)}(\vec{x};\mathcal{W})$ is also impractical for even modest values of $j$. In Appendix~\ref{app:decomp_procedure}, we describe an alternative procedure that efficiently contracts the TR and TTN component tensors in a manner that ultimately yields the same output as the standard contraction, but also separates out the various $\vec{d}^{\ (j)}(\vec{x};\mathcal{W})$ contributions.

In this section, we carry out these interaction decompositions on TR and TTN models that were trained to classify digits from the MNIST~\cite{MNIST} and Fashion MNIST~\cite{Xiao_Rasul_Vollgraf_2017} datasets. These datasets have been widely used to evaluate tensor network models in the literature, and thus serve as reasonable benchmarks for our analysis. Given that the number of operations needed for a full interaction decomposition can scale quadratically with the number of features (see Appendix~\ref{app:decomp_procedure}), we resized each image from $28 \times 28$ pixels to $8 \times 8$ pixels in order to reduce the computational burden of the tests. The grayscale pixels were also normalized to floating-point values on the range $[-0.5, 0.5]$ to improve the numerical stability of the networks. The bond dimension of the TR and TTN models was set to 20, providing them with sufficient representational power without excessive overfitting. The regression output $\vec{f}(\vec{x};\mathcal{W})$ was fit against one-hot encodings of the digit labels and optimized using gradient descent with a mean squared error loss function. During training the networks were contracted normally, with the interaction decomposition being performed at the end using the test dataset.

To begin our analysis, we focus first on the magnitudes of the different \sloppy{$\vec{d}^{\ (j)}(\vec{x};\mathcal{W})$} contributions. To produce a single magnitude for each degree, we computed the L1 norm of $\vec{d}^{\ (j)}(\vec{x};\mathcal{W})$ for each image in the test dataset, and then averaged over the set. Figure~\ref{fig:order_mag} shows the resulting magnitudes for ten TR models and ten TTN models, all trained using the same hyperparameters but with different initial values for the tensor elements. Across both datasets the TR and TTN plots show a similar pattern, with the degree magnitudes starting at approximately $10^{-1}$ for $j=0$ and then growing steadily to some maximum value before declining again at larger $j$. The size and location of the peak varies significantly between the MNIST and Fashion MNIST models, with the MNIST models peaking from 0.1 to 1 at around degrees 10 to 15 while the Fashion MNIST models peak from $10^2$ to $10^4$  at around degrees 17 to 23. After the peak, the magnitudes begin to drop off precipitously, with interaction degrees greater then 45 typically having contributions orders of magnitude smaller than those from degrees before the peak. The inset plots of Figure~\ref{fig:order_mag} show that there is a significant amount of variation between individual models of a given network type and dataset, with some models having magnitudes 10 or even 100 times larger than others.

\begin{figure}
    \centering
    \includegraphics[width=\textwidth]{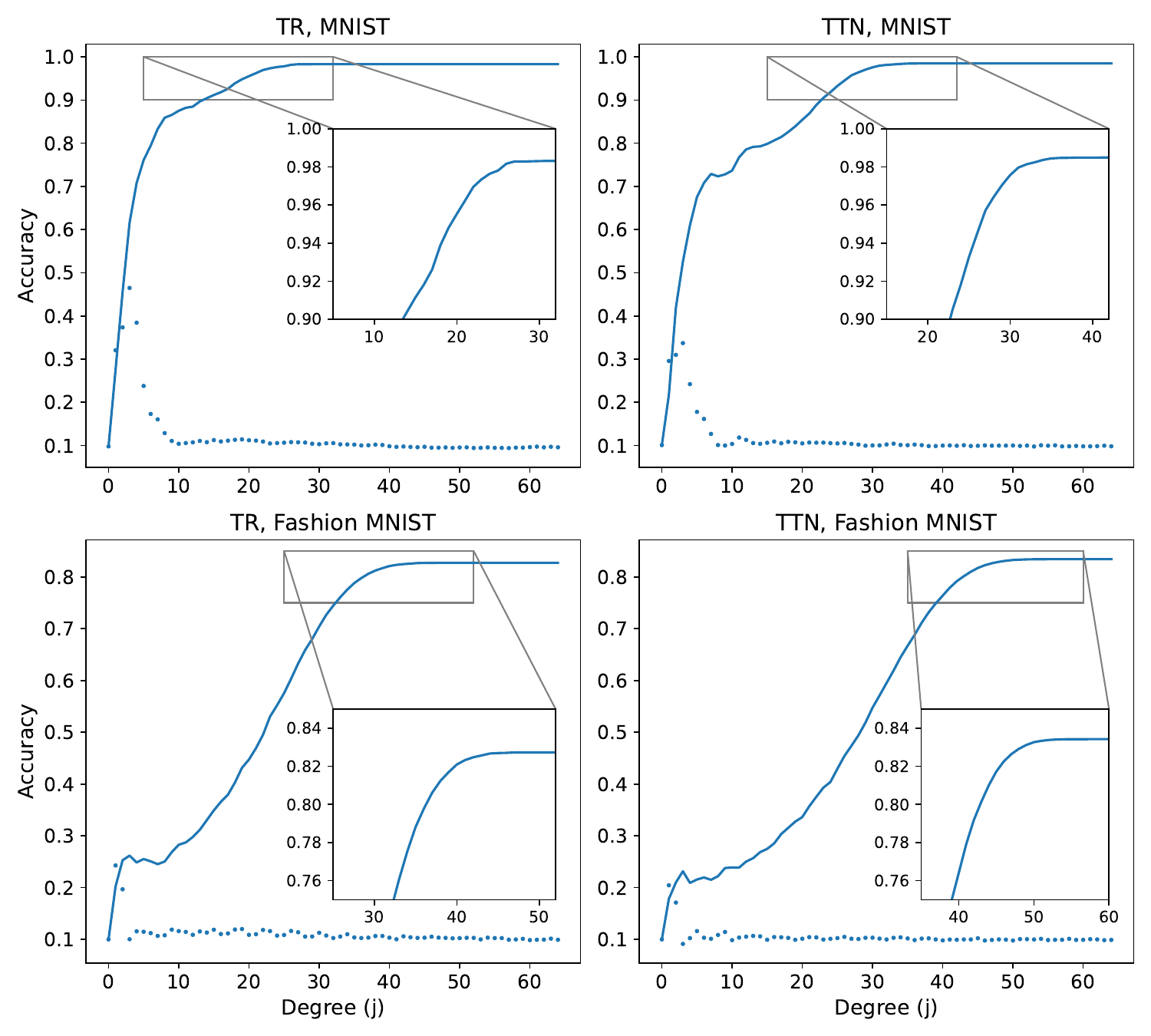}
    \caption{Plots of the average classification accuracy for the TR models and TTN models, trained using all interaction degrees, as a function of interaction degree $j$, using the MNIST and Fashion MNIST test datasets. The scatter plots show the accuracy of each $\vec{d}^{\ (j)}(\vec{x};\mathcal{W})$ term individually, while the solid line shows the accuracy of the sum of contributions from all degrees less than or equal to its position on the x-axis. On MNIST (Fashion MNIST), the cumulative accuracies of the TR and TTN models are equal to $98.31\%$ ($82.73\%$) and $98.49\%$ ($83.43\%$) respectively when all degrees are included, with the performance plateauing at degree 31 (44) for the TRs and 38 (54) for the TTNs. The accuracies of the individual contributions are all very low, with the vast majority of interaction degrees having almost no independent classification ability.}
    \label{fig:accuracies}
\end{figure}

A significant limitation of the magnitude analysis from Figure~\ref{fig:order_mag} is that it can be difficult to assess the true importance of a degree contribution using only its average magnitude. Indeed, even if a set of degrees all have small individual magnitudes, their cumulative effect on the output may still be important. To better assess the ``usefulness'' of the degree contributions, we computed the accuracy of the TR and TTN classifiers as a function of interaction degree, both individually and cumulatively. Figure~\ref{fig:accuracies} shows these accuracies after averaging over the models of each network type. The cumulative accuracy (shown using a solid line) of degree $j$ denotes the accuracy of the output generated by the sum of all degree contributions less than or equal to $j$, while the individual accuracy of degree $j$ (shown as a point on the scatter plot) gives the performance of $\vec{d}^{\ (j)}(\vec{x};\mathcal{W})$ alone. The dimension of the expanded feature space corresponding to each data point is determined by Eq.~\eqref{eq:D_dim}. 

From the plots of cumulative accuracy, we can see that the average performance of both the TR and TTN networks plateaus at slightly over $98\%$ on MNIST ($98.31\%$ for the TRs and $98.47\%$ for the TTNs when all degrees are included), which is consistent with prior work~\cite{Stoudenmire_Schwab_2016}\cite{Stoudenmire_2018}\cite{Efthymiou_Hidary_Leichenauer_2019}. On Fashion MNIST the accuracies are signficantly lower, at $82.73\%$ for the TR and $83.43\%$ for the TTN. \footnote{Tensor networks can achieve significantly higher accuracies on 28 x 28 Fashion MNIST \cite{Glasser_Pancotti_Cirac_2018}\cite{Stoudenmire_2018}\cite{Chen_Pan_Dong_2021}, but the decrease in performance at 8 x 8 is much more severe than for standard MNIST, due to the greater image complexity. For comparison to more state-of-the-art methods, an Inception convolutional network~\cite{Szegedy_Wei_Liu_Yangqing_Jia_Sermanet_Reed_Anguelov_Erhan_Vanhoucke_Rabinovich_2015} can achieve accuracies of 99.09\% on $8 \times 8$ MNIST and 86.5\% on $8 \times 8$ Fashion MNIST (see Appendix~\ref{app:models})} These final accuracy values are of less significance to us than the interaction degree at which the curve flattens. On MNIST (Fashion MNIST) this occurs at approximately $j = 31$ (44) for the TRs, and at $j = 38$ (54) for the TTNs. Looking back at the magnitudes from Figure~\ref{fig:order_mag}, this indicates that even contributions on the order of $10^{-3}$ can still improve the performance of the classifier. Interestingly, all four of the accuracy curves show a temporary flattening before degree 10, followed by a second upward rise. This effect is least visible on the TR MNIST curve and most visible on the two Fashion MNIST curves, with the latter pair of curves seeing most of their accuracy gains after degree 10.

Based on the accuracies of the individual contributions $\vec{d}^{\ (j)}(\vec{x};\mathcal{W})$, which are shown in Figure~\ref{fig:accuracies} using scatter plots, it is clear that only the first few interaction degrees are having their coefficients optimized such that they can classify images independently. The remaining contributions, which constitute the vast majority of regressors, have accuracies close to $10 \%$ and therefore do not separate the different digit classes to any appreciable extent when used in isolation. This suggests that the higher-degree $\vec{d}^{\ (j)}(\vec{x};\mathcal{W})$ have been trained essentially to correct or finesse the cumulative output from the lower degrees, since the cumulative accuracy continues to increase as their outputs are incorporated. This trend is particularly marked for the Fashion MNIST models, where only degrees 1 and 2 have accuracies above $12\%$. Indeed, plots C and D from Figure~\ref{fig:order_mag} show that many of the regressors in these models are being used to cancel out the large magnitudes of the intermediate degrees, since the final regression output needs to be roughly in the range [0, 1] to achieve a reasonable loss value.

\subsection{Interaction Decompositions as Regression Models}\label{sec:decomp_regression}

In Sec.~\ref{sec:tr_ttn_decomp}, we used the interaction decomposition as a tool to analyze tensor network models that had been trained using standard methods. As a result, the parameters of each model were optimized under the assumption that every interaction degree would contribute to the final output, without any truncation or isolation. This offers the greatest flexibility to the model in principle, but it can also obscure the potential success that a single degree or subset of degrees might have had if the parameters of the network had been optimized to improve their performance specifically.

In light of this fact, we propose a new type of tensor network model called the $\mathcal{D}$-degree tensor network. In these models, only interaction degrees in the set $\mathcal{D}$ are used to construct the regression output, such that 
\begin{equation}
    \vec{f}(\vec{x};\mathcal{W}) = \sum_{j \in \mathcal{D}} \vec{d}^{\ (j)}(\vec{x};\mathcal{W}).
\end{equation}
Comparing this expression to the full interaction decomposition given in Eq.~\eqref{eq:degree_decomp}, it is clear that if $\mathcal{D}$ is the set of all interaction degrees (i.e. if $\mathcal{D} = \{0, 1, ..., m\}$), then the corresponding $\mathcal{D}$-degree network is equivalent to a standard tensor network with the same structure. However, we will focus our attention on models where $\mathcal{D}$ contains only a fraction of the $m + 1$ possible interaction degrees. By restricting the regression in this manner, we are effectively inducing sparsity in the weight tensor $W$ by zeroing the coefficients for all interaction degrees not included in $\mathcal{D}$. However, unlike in the case of sparse neural networks~\cite{Srinivas_Subramanya_Babu_2017}\cite{Liu_Wang_Foroosh_Tappen_Penksy_2015}, this sparsity does not necessarily lead to a reduction in the number of trainable parameters or to an improvement in the computational overhead. Instead, the sparsity leads to a simplification in the structure of the regression function, which can yield a model that is more easily interpretable while still achieving the same level of performance.

Using the decomposition procedure described in Appendix~\ref{app:decomp_procedure}, it is possible to efficiently train $\mathcal{D}$-degree models on the same regression tasks used in Sec.~\ref{sec:tr_ttn_decomp}, and thus compare their classification accuracies with those shown in Figure~\ref{fig:accuracies}. For our tests, we selected $\mathcal{D}$-degree models that fell into two categories: the \textit{cumulative-j} models, in which all degrees less than or equal to $j$ are included in the output, and the \textit{degree-j} models, in which the output is simply the contribution from the $j$th degree:
\begin{equation}\label{eq:degree_models}
    \text{Cumulative-}j:\ \vec{f}(\vec{x};\mathcal{W}) = \sum^j_{j'=0} \vec{d}^{\ (j')}(\vec{x};\mathcal{W}), \quad \quad \text{Degree-}j: \  \vec{f}(\vec{x};\mathcal{W}) = \vec{d}^{\ (j)}(\vec{x};\mathcal{W}).
\end{equation}
These two groups describe only a small portion of the $2^{m+1} - 1$ possible $\mathcal{D}$-degree models, but they will allow us to easily compare our results with the plots in Figure~\ref{fig:accuracies}. For our numerical tests, we trained the models on $8 \times 8$ images from the MNIST and Fashion MNIST datasets prepared in the same manner described in Sec.~\ref{sec:tr_ttn_decomp}. The $\mathcal{D}$-degree models take somewhat longer to train than standard tensor network models due to the added complexity of the interaction decomposition, but their times are still on par with those of neural network models (see Appendix~\ref{app:models}). 

Figure~\ref{fig:trained} shows average accuracies of the cumulative-$j$ (blue plots) and degree-$j$ (orange plots) models as a function of $j$, with each data point representing an average across ten models. These averages are plotted alongside the cumulative data from Figure~\ref{fig:accuracies}, which shows the performance of the full tensor network models as a reference. We emphasize that the cumulative-$j$ and degree-$j$ curves in Figure~\ref{fig:trained} are computed in precisely the same manner as the line and scatter plots from Figure~\ref{fig:accuracies}, except that the models which generated Figure~\ref{fig:accuracies} were trained using all of the interaction degrees rather than just the specific subset being plotted. The plots for the $\mathcal{D}$-degree models omit results for $j = 0$, since those models contain only the bias term and thus predict the same digit for every image. The data used to generate these plots is given in Appendix~\ref{app:tables}.

\begin{figure}
    \centering
    \includegraphics[width=\textwidth]{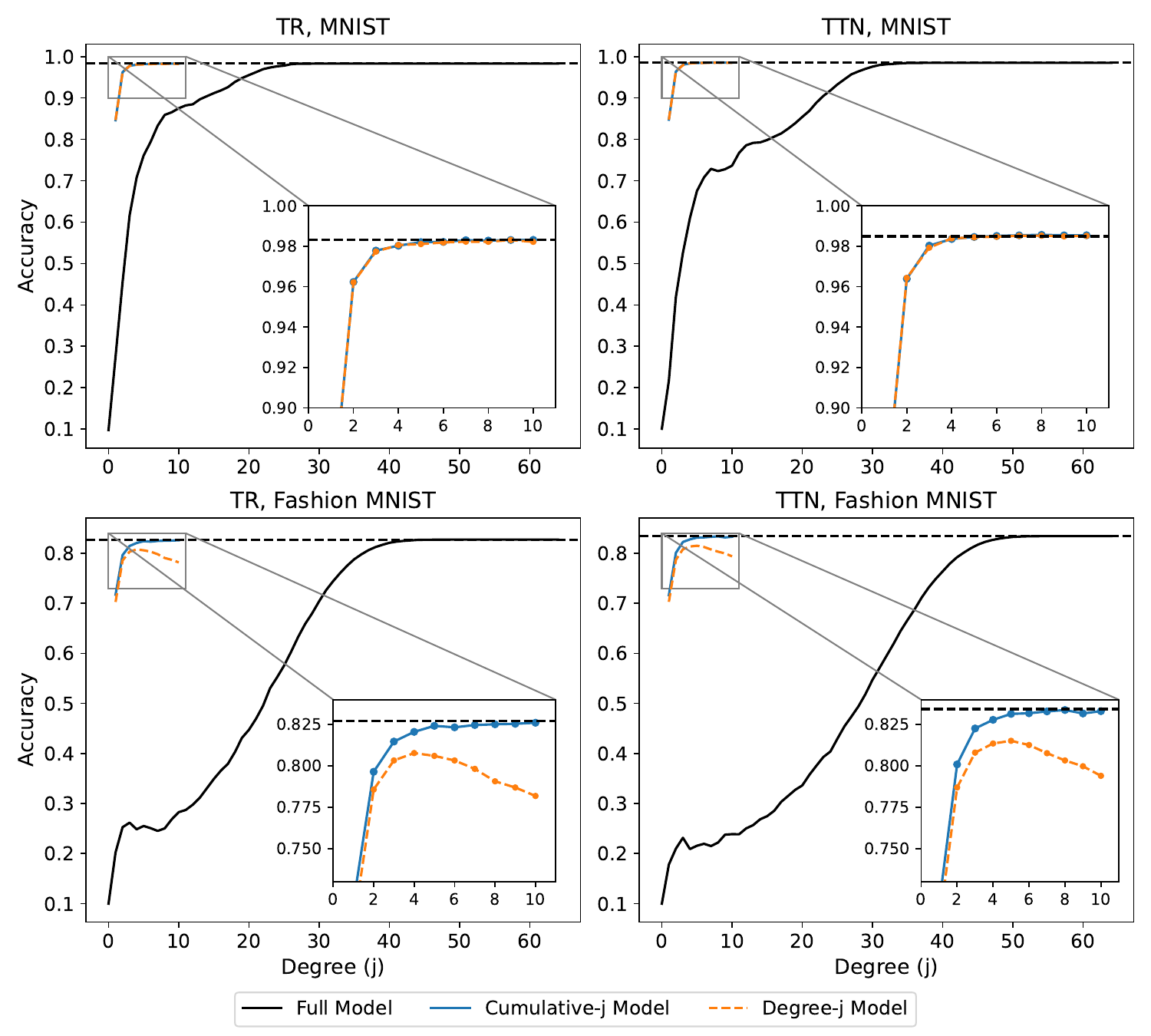}
    \caption{Plots of average accuracy on the MNIST and Fashion MNIST test datasets for $\mathcal{D}$-degree models parameterized by TRs and TTNs, with ten models being averaged for each degree. The solid black line indicates the cumulative accuracy of the standard tensor network models analyzed in Sec.~\ref{sec:tr_ttn_decomp}, as plotted in Figure~\ref{fig:accuracies}, with the dashed line showing the accuracy when all interaction degrees are included. The degree-$j$ models generally demonstrate worse performance than the cumulative-$j$ models, though the difference is very small on MNIST. The cumulative-$j$ models were able to closely match the accuracies of the full models on both datasets, and even slightly exceed their performance on MNIST.}
    \label{fig:trained}
\end{figure}

The first trend to observe from Figure~\ref{fig:trained} is that both the cumulative-$j$ and degree-$j$ models have accuracies that are significantly greater than the corresponding cumulative accuracy at degree $j$ from the full tensor network models. This performance gap is notable, because it implies that the standard models are utilizing the feature-product regressors in a highly inefficient manner. For MNIST in particular, the degree-$j$ classifiers with $j > 3$ were able to perform within $0.5\%$ of the full model. As a comparison, the cumulative MNIST accuracy of the regular TR and TTN models using degrees $0\text{ -- }4$ is only $71\%$ and $61\%$ respectively. The disparity is even larger when looking at the single-degree accuracies from Figure~\ref{fig:accuracies}, which show that most individual $\vec{d}^{\ (j)}(\vec{x};\mathcal{W})$ were unable to classify images at all when trained as part of a full tensor network model. When those degree contributions were optimized directly, however, they were able to perform classification with more than $98\%$ accuracy. The degree-$j$ models did not perform as well on Fashion MNIST, though they still achieved accuracies that were vastly higher than the corresponding cumulative accuracies from Figure \ref{fig:accuracies}. The cumulative-$j$ models, on the other hand, were able to closely match the performance of the standard models, with the cumulative-10 TR and cumulative-8 TTN models coming within $0.1 \%$ of their full-degree counterparts.

The dashed horizontal lines in Figure~\ref{fig:trained} mark the accuracy of the full tensor network models when every degree contribution is included. Using these values as a benchmark, we can see that several of the cumulative-$j$ TTN models ($6 \leq j \leq 10$) and degree-$j$ TTN models ($7 \leq j \leq 10$) actually \textit{outperformed} the corresponding full model on the MNIST dataset. This is a counter-intuitive result, as it suggests that regressing on all of the interaction degrees can actually yield slightly worse results than regressing on only a small subset of them. A cumulative-8 TTN model, for example, uses only one-billionth of the feature products contained within the data tensor $X$, yet achieves an average accuracy roughly $0.1\%$ higher than a TTN model which has access to all of $X$.

Finally, we note that a comparison can be made between the performance of these $\mathcal{D}$-degree network models, which constrain the feature product coefficients to all be generated by the same low-rank tensor network, and a more general multilinear regression model in which every coefficient can be determined arbitrarily. In Appendix~\ref{app:models}, we give results for this type of unconstrained regression on features products up to degree 4, which shows that the cumulative-$j$ models achieve accuracies very near the arbitrary models of degree $j$, and can outperform them for larger values of $j$ even when the tensor network models contain fewer trainable parameters. This demonstrates the utility of incorporating more interactions (up to a point), since constrained regression on higher-degree feature products is more effective than unconstrained regression on lower-degree feature products.

\section{Discussion}\label{sec:discussion}

The exponential feature space induced by the transformation in Eq.~\eqref{eq:data_tensor} lies at the heart of tensor network regression, and there is no doubt that these models utilize it to achieve a level of performance that far exceeds standard linear regression. That said, it is easy to feel incredulous toward the idea that tensor network models, or indeed any regression model, could truly make use of the $2^{64}$ different regressors that are generated from an $8 \times 8$ image. The goal of our work here has been to develop the interaction decomposition as a tool to test this claim, and then apply it to tensor network models under a pair of standard machine learning tasks. By evaluating the magnitudes and accuracies of the different interaction degrees, we can begin to draw conclusions about how effectively the exponential space is being utilized.

To this end, our results from Sec.~\ref{sec:tr_ttn_decomp} show that more than half of the interaction degrees contributed meaningfully to the output of the classifiers, with the Fashion MNIST TTN models in particular using up to degree 50. In the language of Sec.~\ref{sec:degree_subspaces}, this indicates that the tensor network models are utilizing a portion of the expanded feature space $\mathbb{X}$ that has a dimension on the order of $10^{19}$, which can be computed by summing Eq.~\eqref{eq:degree_dim} across all significant degrees. It is important to note, however, that Figure~\ref{fig:accuracies} only shows the change in accuracy for the $j$th interaction degree when the entirety of subspace $\mathbb{D}^{(j)}$ is incorporated into the regression. It may very well be that the models in Sec.~\ref{sec:tr_ttn_decomp} are utilizing only a small portion of \textit{this} space, and thus the number of relevant feature products could be far smaller than the upper-bound of $10^{19}$. The interaction decomposition cannot separate out different parts of a given degree-$j$ subspace, so future work might look into alternate algorithms that are able to divide up these spaces into meaningful components.

For a given interaction degree, one can ask not only \textit{if} the set of feature products is being utilized by a tensor network model, but also \textit{how well} the model is using them relative to some standard. In Sec.~\ref{sec:decomp_regression} we introduced the $\mathcal{D}$-degree tensor network to serve as this standard, since its parameters could be trained to maximize performance using only a specific subset of interaction degrees. In our tests, the networks were limited to interaction degrees of at most 10, which corresponds to an expanded features space of dimension at most $10^{11}$ as given by Eq.~\eqref{eq:D_dim}. The results shown in Figure~\ref{fig:trained} demonstrate that the full tensor network models are significantly under-utilizing the lower-degree interactions, since the $\mathcal{D}$-degree models are able to achieve accuracies up to 60 percentage points higher when constrained to those same degrees. This under-utilization was especially acute for Fashion MNIST, where the full models only reached cumulative accuracies of $25\%$ for the first ten degrees, despite the fact that the cumulative-10 models had accuracies near $83 \%$. 

More significantly, some $\mathcal{D}$-degree models trained using only the first six interaction degrees were able to achieve accuracies on MNIST that were \textit{greater} than those of models trained using all degrees. While this could simply be due to more overfitting in the full models, it might also point to inherent limitation in the tensor network representation of $W$. We know, for example, that the regression coefficients in a tensor network model are necessarily coupled together by the elements of the component tensors, which may force the model to use suboptimal coefficients for the lower interaction degrees in order to avoid harmful contributions from the higher degrees. Given that detailed, ``under-the-hood'' analyses of these models are possible using methods such as the interaction decomposition introduced here, the existence and nature of this compromise seems like a promising area for further study.

Taken together, the results discussed here support the following two conclusions:
\begin{enumerate}
    \item Common tensor network models are capable of utilizing regressors from a large portion of the expanded feature space generated by the featurization from \cite{Novikov_Trofimov_Oseledets_2016}.
    \item However, a comparable level of performance may also be achieved by regression on a minuscule fraction of that same space.
\end{enumerate}
For those looking to use tensor network models for machine learning, there is cause here for both optimism and caution. While the first conclusion makes it clear that tensor network regression models can incorporate useful information from a wide range of interaction degrees, the second conclusion implies that it is difficult for these models to extract any \textit{unique} information from the higher-degree regressors. In light of this, we believe that the $\mathcal{D}$-degree tensor network models, which have been absent in the literature up to this point, represent a promising approach for tensor network regression. Using these models, it is possible to exploit the representational efficiency of tensor networks while constraining the regression to a reasonable and interpretable set of feature products based on the inherent complexity of the dataset. 

\section*{Acknowledgements}

Funding for this work was provided by the UC Noyce Initiative.

\printbibliography

\appendix

\section{Appendix}\label{sec:appendix}

\subsection{Procedure for the Interaction Decomposition}\label{app:decomp_procedure}

In this section, we describe a procedure that can be used to carry out the interaction decomposition of any tensor network. At its core is a tensor operation that we call the \textit{degree-preserving tensor product}, denoted $\tilde{\otimes}$, which is defined between $m+1$th order tensor $A$ and $n+1$th order tensor $B$ as
\begin{equation}\label{eq:degree_tensor_product}
    (A\ \tilde{\otimes}\ B)_{ji_0...i_{m-1}k_0...k_{n-1}} = \sum_{j_a + j_b = j}A_{j_ai_0...i_{m-1}}B_{j_bk_0...k_{n-1}},
\end{equation}
where the resulting tensor is of order $m + n + 1$ and $0 \leq j \leq \text{max}(j_a) + \text{max}(j_b)$. Note that this operation attaches special significance to the first dimension, which we will hereafter refer to as the \textit{degree index}. As shown in Eq.~\eqref{eq:degree_tensor_product}, the $j$th slice of $A\ \tilde{\otimes}\ B$ along the degree index is given by the sum of tensor products taken between slices of $A$ and of $B$, such that the sum of the degree indices for those slices is equal to $j$. Like the normal tensor product, the degree-preserving tensor product is associative and commutative up to a permutation of the (non-degree) indices, and multilinear in its two arguments. Using this new variation of the tensor product, we can also define a \textit{degree-preserving contraction} in the same manner as Eq.~\eqref{eq:tensor_contraction}, such that the contraction of fourth-order tensors $A$ and $B$ is given by
\begin{equation}
    C_{jklqr} = \sum_{j_a + j_b = j}\sum_i A_{j_akil} B_{j_bqri}.
\end{equation}

The utility of these degree-preserving operations becomes apparent if we alter the featurization in Eq.~\eqref{eq:feat_func} to be
\begin{equation}\label{eq:H_feat}
    H^{(i)}(x_i) = 
    \begin{bmatrix}
        1 & 0 \\
        0 & x_i,
    \end{bmatrix},
\end{equation}
which simply embeds $\vec{h}^{(i)}(x_i)$ along the diagonal of a $2 \times 2$ matrix. Note that the degree index of this tensor matches up with the interaction degree of its non-zero elements, since the first row (index 0) is a constant while the second row (index 1) is $x_i$. This correspondence is maintained by the  degree-preserving tensor product of $H^{(i)}$ and $H^{(k)}$:
\begin{equation}\label{eq:H_tensor_prod}
    H^{(i)}\ \tilde{\otimes}\ H^{(k)} = 
    \begin{bmatrix}
    \ & \ & \ \\
        \begin{bmatrix}
            1 & 0 \\
            0 & 0
        \end{bmatrix}, &
        \begin{bmatrix}
            0 & x_i \\
            x_k & 0
        \end{bmatrix}, &
        \begin{bmatrix}
            0 & 0 \\
            0 & x_ix_k
        \end{bmatrix} \\
        \ & \ & \ 
    \end{bmatrix},
\end{equation}
where the non-zero elements all have an interaction degree equal to their position along the degree index. Since the zero elements do not contribute anything during a tensor contraction, Eq.~\eqref{eq:H_tensor_prod} also indicates that any degree-preserving contraction between $H^{(i)}$ and $H^{(k)}$ would likewise maintain the correspondence between degree index and interaction degree.

Using the degree-preserving tensor product and contraction operations, along with the new featurization maps $H^{(i)}(x_i)$, the interaction decomposition of a tensor network regression model can be carried out using the following procedure:
\begin{enumerate}
    \item Add a degree index of size one (i.e., an index that can only take a value of 0) to each component tensor of the network representing $W$. This increases the order of each tensor by one, but leaves the actual number of elements unchanged. 
    
    \item Construct (implicitly) a modified data tensor $\tilde{X}$ using the mappings from Eq.~\eqref{eq:H_feat}, such that $\tilde{X}(\vec{x}) = H^{(0)}(x_0)\ \tilde{\otimes}\ H^{(1)}(x_1)\ \tilde{\otimes}\ ... \ \tilde{\otimes}\ H^{(m-1)}(x_{m-1})$.
    
    \item Use degree-preserving contraction operations to contract $\tilde{X}$ with the tensor network, following whichever efficient contraction scheme is appropriate for the network architecture of the model.
    
    \item If the decomposition is being used to contract a $\mathcal{D}$-degree network, then the degree index of all intermediate tensors can be be truncated to the largest degree in $\mathcal{D}$.
\end{enumerate}
Since the contraction of the network is done using degree-preserving contractions, the contributions from each interaction degree are kept separate throughout the entire process. The final output of the interaction decomposition (without truncation) is a second-order tensor of the form
\begin{equation}
    F(\vec{x};\mathcal{W}) =
    \begin{bmatrix}
        \vec{d}^{\ (0)}(\vec{x};\mathcal{W}), & \vec{d}^{\ (1)}(\vec{x};\mathcal{W}), & ..., & \vec{d}^{\ (m)}(\vec{x};\mathcal{W}) 
    \end{bmatrix},
\end{equation}
where $\vec{d}^{\ (j)}(\vec{x};\mathcal{W})$ is the degree-$j$ contribution to the combined regression output $\vec{f}(\vec{x};\mathcal{W})$.
The computational cost of the procedure described above is best understood in terms of how much additional complexity it adds on top of a standard contraction of the network. This complexity comes from two sources: larger intermediate tensors due to the addition of the degree index, and an extra sum over the degree index that is present in the degree-preserving tensor product from Eq.~\eqref{eq:degree_tensor_product}. The first contribution is easy to characterize, since adding a degree index simply increases the size of the original tensors by a factor that is on the order of the maximum interaction degree $j_{max}$ in the decomposition. The second contribution is more subtle, since the number of terms in the tensor-product sum depends on the relative sizes of the degree indices of the two inputs. Consider again the tensor product between $A$ and $B$ from Eq.~\eqref{eq:degree_tensor_product}, and let $\bar{j}_a$ and $\bar{j}_b$ be the largest value of the degree index for $A$ and $B$ respectively, with $\bar{j} = \bar{j}_a + \bar{j}_b$ and $\bar{j}_a \leq \bar{j}_b$. Then it it can be shown that the number of terms $s$ needed to generate all $\bar{j} + 1$ slices of $A\ \tilde{\otimes}\ B$ is given by
\begin{equation}
    s = (\bar{j}_a + 1)(\bar{j}_b + 1),
\end{equation}
which scales as $\mathcal{O}(\bar{j}_a\bar{j}_b)$. This means that, for a fixed $\bar{j}$, the value of $s$ can range from a minimum of $\bar{j} + 1$ if $\bar{j}_a = 0$ to a maximum of $\frac{1}{4}(\bar{j})^2 + \bar{j} + 1$ for the fully symmetric case when $\bar{j}_a = \bar{j}_b = \frac{1}{2}\bar{j}$. Given that the last contractions in the interaction decomposition will have $\bar{j}$ on the order of $j_{max}$, this means that the most complex degree-preserving tensor products can either scale as $\mathcal{O}(j^2_{max})$ or $\mathcal{O}(j_{max})$, depending on the amount of symmetry between the two input tensors. The fact that more symmetric contraction schemes can lead to worse scaling (quadratic rather than linear in $j_{max}$) is an interesting property of this method, although the use of such schemes may still be desirable due to other computational advantages.

\subsection{Tabulation of \texorpdfstring{$\mathcal{D}$}{D}-degree Model Performance}\label{app:tables}

The following two tables show the results of the numerical tests described in Sec.~\ref{sec:decomp_regression} and plotted in Figure~\ref{fig:trained}, along with the relevant cumulative accuracy values for the full models from Figure~\ref{fig:accuracies}. Table~\ref{tab:mnist} gives the accuracies for MNIST, while Table~\ref{tab:fashion} provides them for Fashion MNIST. Each value represents the average percent test accuracy across ten different initializations of the given model type, with the standard error of the last digit shown in parentheses. For the cumulative-$j$ and degree-$j$ models the column label denotes the value of $j$, while for the full models they denote the cumulative accuracy of the output up to the $j$th interaction degree.

\begin{table}
    \centering
    \resizebox{\textwidth}{!}{
        \begin{tabular}{c|c|c|c|c|c|c|c|c|c|c|c}
         & 1 & 2 & 3 & 4 & 5 & 6 & 7 & 8 & 9 & 10 & 64 \\ \hline
          Full TR & $27.5(5)$ & $46(1)$ & $62(1)$ & $70.8(8)$ & $76(1)$ & $79(2)$ & $83(3)$ & $86(3)$ & $87(3)$ & $88(3)$ & $98.31(2)$ \\ Cumulative TR &$84.6(1)$ & $96.23(3)$ & $97.79(3)$ & $98.03(3)$ & $98.21(2)$ & $98.21(3)$ & $98.30(4)$ & $98.28(3)$ & $98.31(4)$ & $98.31(2)$ & - \\ Degree TR &$84.67(6)$ & $96.22(2)$ & $97.74(3)$ & $98.06(3)$ & $98.11(2)$ & $98.19(3)$ & $98.23(2)$ & $98.22(3)$ & $98.31(3)$ & $98.22(3)$ & - \\ Full TTN & $21.7(6)$ & $42(1)$ & $53(1)$ & $61(1)$ & $68(1)$ & $71(2)$ & $73(2)$ & $72(2)$ & $73(2)$ & $74(3)$ & $98.49(3)$ \\ Cumulative TTN &$84.76(7)$ & $96.39(2)$ & $98.03(2)$ & $98.36(2)$ & $98.46(1)$ & $98.51(2)$ & $98.54(3)$ & $98.57(1)$ & $98.54(1)$ & $98.54(1)$ & - \\ Degree TTN &$84.76(8)$ & $96.44(2)$ & $97.93(2)$ & $98.39(2)$ & $98.44(3)$ & $98.47(2)$ & $98.52(3)$ & $98.53(2)$ & $98.49(2)$ & $98.51(1)$ & - 
        \end{tabular}}
    \caption{Table of average accuracy vs degree for the six different model types on MNIST, for Figure~\ref{fig:trained}.}
    \label{tab:mnist}
    \vspace{0.5in}
    \resizebox{\textwidth}{!}{
        \begin{tabular}{c|c|c|c|c|c|c|c|c|c|c|c}
              & 1 & 2 & 3 & 4 & 5 & 6 & 7 & 8 & 9 & 10 & 64 \\ \hline
             Full TR & $20.2(4)$ & $25.3(5)$ & $26(1)$ & $25(1)$ & $25.5(9)$ & $25(1)$ & $25(1)$ & $25(1)$ & $27(1)$ & $28(1)$ & $82.73(9)$ \\ Cumulative TR &$71.73(6)$ & $79.64(7)$ & $81.47(5)$ & $82.05(8)$ & $82.42(7)$ & $82.3(1)$ & $82.47(6)$ & $82.51(7)$ & $82.54(9)$ & $82.60(7)$ & - \\ Degree TR &$70.27(8)$ & $78.57(5)$ & $80.33(8)$ & $80.77(7)$ & $80.60(6)$ & $80.3(1)$ & $79.82(7)$ & $79.07(6)$ & $78.7(1)$ & $78.18(9)$ & - \\ Full TTN & $17.8(3)$ & $21.0(3)$ & $23.2(8)$ & $21(1)$ & $22(1)$ & $22(1)$ & $22(1)$ & $22(1)$ & $24(2)$ & $24(2)$ & $83.43(6)$ \\ Cumulative TTN &$71.63(7)$ & $80.09(6)$ & $82.26(7)$ & $82.78(6)$ & $83.14(6)$ & $83.18(7)$ & $83.29(7)$ & $83.37(6)$ & $83.17(9)$ & $83.31(4)$ & - \\ Degree TTN &$70.30(4)$ & $78.69(6)$ & $80.80(5)$ & $81.35(6)$ & $81.51(5)$ & $81.26(8)$ & $80.76(8)$ & $80.33(9)$ & $80.0(1)$ & $79.4(1)$ & -  
            \end{tabular}}
    \caption{Table of average accuracy vs degree for the six different model types on Fashion MNIST, for Figure~\ref{fig:trained}.}
    \label{tab:fashion}
\end{table}

\subsection{Regression Model Comparisons}\label{app:models}

In Table~\ref{tab:models}, we compare our TR and TTN models with several low-order multilinear models and a deep learning model, in terms of their number of trainable parameters, computation time per epoch, and average accuracies on the $8 \times 8$ image datasets. The linear, bilinear, trilinear, and tetralinear regression models perform unconstrained regression on feature products of degree less than or equal to 1, 2, 3, and 4 respectively, which are the same regressors used by the cumulative-$j$ models for $1 \leq j \leq 4$. By ``unconstrained'',  we mean that the coefficients for each feature product can be set arbitrarily rather than being generated by a low-rank tensor network. To offer a comparison with state-of-the-art neural network algorithms, we also provide the corresponding numbers for a convolutional neural network (CNN) model based on the Inception~\cite{Szegedy_Wei_Liu_Yangqing_Jia_Sermanet_Reed_Anguelov_Erhan_Vanhoucke_Rabinovich_2015} architecture, which contains the most parameters and achieves the best performance on both datasets.

\begin{table}[h]
    \centering
    \begin{tabular}{c|c|c|c|c}
         & Parameters & \makecell{Seconds \\ per epoch} & \makecell{MNIST \\ accuracy} & \makecell{Fashion MNIST \\ accuracy} \\ \hline
         Linear & 65 & 2 & 84.7(1) & 71.35(8)
         \\
         Bilinear & 2,081 & 2 & 96.47(1) & 80.20(6)
         \\
         Trilinear & 43,745 & 3 & 98.13(2) & 82.32(7)
         \\
         Tetralinear & 679,121 & 11 & 98.46(1) & 82.33(3)
         \\
         Full TR & 51,200 & 2 & 98.31(2) & 82.73(9)
         \\
         Full TTN & 250,560 & 3 & 98.49(3) & 83.43(6)
         \\ 
         Cumulative-10 TR & 51,200 & 29 & 98.31(2) & 82.60(7)
         \\
         Cumulative-8 TTN & 250,560 & 18 & 98.57(1) & 83.37(6)
         \\
         Inception CNN & 1,196,530 & 23 & 99.27(3) & 86.64(9)
    \end{tabular}
    \caption{Table of parameter number, seconds of computation per epoch (with batch size of 64), and average classification accuracies on MNIST and Fashion MNIST for various regression models. The averages were computed across ten different initializations, with the standard error of the last digit given in parentheses.}
    \label{tab:models}
\end{table}

\end{document}